\documentclass[a4paper]{llncs}

\usepackage[latin1]{inputenc}
\usepackage{amsmath}
\usepackage{amssymb}
\usepackage[amsmath,thmmarks]{ntheorem}
\usepackage{color}

\def\CPP{\leavevmode\textrm{\hbox{C\hskip
-0.1ex\raise 0.5ex\hbox{\tiny ++}}}}

\newcommand{\Var}{\ensuremath{\mathit{Var}}}
\newcommand{\Val}{\ensuremath{\mathit{Val}}}
\newcommand{\Asn}{\ensuremath{\mathit{Asn}}}
\newcommand{\Dom}{\ensuremath{\mathit{Dom}}}
\newcommand{\Con}{\ensuremath{\mathit{Con}}}
\newcommand{\dom}[1]{\ensuremath{\operatorname{dom}(#1)}}
\newcommand{\dominj}{dom injective}
\newcommand{\conv}{\ensuremath{\operatorname{conv}}}
\newcommand{\vars}{\ensuremath{\mathrm{vars}}}
\newcommand{\stronger}{\subseteq}
\newcommand{\strstronger}{\subset}
\newcommand{\restrict}[2]{\ensuremath{{#1}_{|#2}}}
\newcommand{\varphihat}{\ensuremath{\widehat{\varphi}}}
\newcommand{\varphihatprime}{\ensuremath{\widehat{\varphi'}}}

\newcommand{\mimpl}{\ensuremath{\Rightarrow}}

\newcommand{\lequiv}{\ensuremath{\leftrightarrow}}
\newcommand{\fun}[2]{\ensuremath{#1\rightarrow #2}}
\newcommand{\Power}[1]{\ensuremath{2^{#1}}}
\newcommand{\setc}[2]{\{#1\;|\;#2\}}
\newcommand{\xor}{\oplus}

\newcommand{\ZZ}{\ensuremath{\mathbb Z}}
\newcommand{\RR}{\ensuremath{\mathbb R}}
\newcommand{\DD}{\ensuremath{\mathcal D}}

\newcommand{\boundsd}{\ensuremath{\operatorname{bounds}(\DD)}}
\newcommand{\boundsz}{\ensuremath{\operatorname{bounds}(\ZZ)}}
\newcommand{\boundsr}{\ensuremath{\operatorname{bounds}(\RR)}}

\newcommand{\pparagraph}[1]{\paragraph{\normalfont\textbf{#1.}}}

\newcommand{\xgetmin}[1]{\underline{#1}}
\newcommand{\xgetmax}[1]{\overline{#1}}

\newcommand{\xadjmin}[2]{\underline{#1}\gets #2}
\newcommand{\xadjmax}[2]{\overline{#1}\gets #2}

\newcommand{\reify}[2]{\left(#1\right)\Leftrightarrow #2}
\newcommand{\distinct}[1]{\operatorname{alldifferent}(#1)}
\newcommand{\element}[2]{#1\left[#2\right]}

\theoremstyle{plain}
\theoremheaderfont{\bfseries}
\theorembodyfont{\upshape}
\theoremsymbol{\ensuremath{_\Box}}
\theoremseparator{.}
\renewtheorem{theorem}{Theorem}
\renewtheorem{lemma}{Lemma}
\theoremstyle{nonumberplain}
\renewtheorem{example}{Example}

\theoremstyle{nonumberplain}
\theoremheaderfont{\slshape}
\theorembodyfont{\upshape}
\theoremsymbol{\rule{.8ex}{.8ex}}
\theoremseparator{.}
\renewtheorem{proof}{Proof}

\begin{document}

\sloppy

\title{Perfect Derived Propagators}
\author{Christian Schulte\inst{1} \and Guido Tack\inst{2}}
\institute{
{ICT, KTH - Royal Institute of Technology, Sweden, 
 \texttt{cschulte@kth.se}}
\and
{PS Lab, Saarland University, Saarbrücken, Germany,
 \texttt{tack@ps.uni-sb.de}}}

\maketitle

\begin{abstract} 
  When implementing a propagator for a constraint, one must decide
  about variants: When implementing $\min$, should one also implement
  $\max$?  Should one implement linear equations both with and without
  coefficients?  Constraint variants are ubiquitous: implementing them
  requires considerable (if not prohibitive) effort and decreases
  maintainability, but will deliver better performance.
  
  This paper shows how to use variable views, previously introduced
  for an implementation architecture, to derive \emph{perfect}
  propagator variants.  A model for views and derived propagators is
  introduced. Derived propagators are proved to be indeed perfect in
  that they inherit essential properties such as correctness and
  domain and bounds consistency. Techniques for systematically
  deriving propagators such as transformation, generalization,
  specialization, and channeling are developed for several variable
  domains. We evaluate the massive impact of derived
  propagators. Without derived propagators, Gecode would require
  $140\,000$ rather than $40\,000$ lines of code for propagators.
\end{abstract}

\section{Introduction}
\label{sec:introduction}

When implementing a propagator for a constraint, one typically
needs to decide whether to also implement some of its variants.
For example, when implementing a propagator for $\max_{i=1}^n
x_i=y$, should one also implement $\min_{i=1}^n x_i=y$? When
implementing the linear equation $\sum_{i=1}^n a_i x_i=c$ for
integer variables $x_i$ and integers $a_i$ and $c$, should one
also implement $\sum_{i=1}^n x_i=c$ for better performance? When
implementing the reified linear equation $\reify{\sum_{i=1}^n
  x_i=c}{b}$, should one also implement its almost identical
algebraic variant $\reify{\sum_{i=1}^n x_i\neq c}{b}$?

Implementing inflates code and documentation. Not implementing
increases space and runtime: by using more general propagators
or by decomposing into several other constraints.  Worse, given
the potential code explosion, one may be able to only implement
some variants (say, minimum and maximum). Other variants
important for performance (say, minimum and maximum for two
variables) may be infeasible due to excessive programming
and maintenance effort.

Here, we follow a third approach: we derive propagators from already existing
propagators using variable views. In~\cite{SchulteTack:Advances:2006}, we
introduced an implementation architecture for variable views to reuse generic
propagators without performance penalty. This architecture has been
implemented in Gecode~\cite{Gecode:2008}, and is in fact essential for the
system, as it saves approximately $100\,000$ lines of code. Due to the massive
use of views in Gecode, it is vital to develop a model that allows us to prove
that derived propagators have the desired properties.

In this paper, we argue that propagators that are derived using variable views
are indeed \emph{perfect}: they are not only perfect for performance, we prove
that they inherit all essential properties such as correctness and
completeness from their original propagator.

Last but not least, we show common techniques for deriving propagators with
views and demonstrate their wide applicability. In Gecode, every propagator
implementation is reused $3.6$ times on average. Without views, Gecode would
feature $140\,000$ rather than $40\,000$ lines of propagator implementation to
be written, tested, and maintained.

\pparagraph{Variable views}
Consider a bounds consistent propagator for $\max(x,y)=z$. Assume
that $\xgetmax{x}$ ($\xgetmin{x}$) returns the maximum (minimum)
of the finite domain variable $x$, whereas $\xadjmax{x}{n}$
($\xadjmin{x}{n}$) adjusts the maximum (minimum) value of $x$ to
$\min(\xgetmax{x},n)$ ($\max(\xgetmin{x},n)$), only taking variable bounds into account. The propagator is
implemented by performing the following operations on its
variables:
$$
\xadjmax{x}{\xgetmax{z}}\qquad \xadjmax{y}{\xgetmax{z}}\qquad
\xadjmax{z}{\max(\xgetmax{x},\xgetmax{y})}\qquad
\xadjmin{z}{\max(\xgetmin{x},\xgetmin{y})}
$$

Given three more propagators for $x'=-x$, $y'=-y$, and $z'=-z$, we could
propagate the constraint $\min(x',y')=z'$. In contrast to this
\emph{decomposition}, we propose to use generic propagators that perform
operations on views rather than variables. Views provide the same interface
(set of operations) as variables while enabling additional transformations.
For example, an operation on a minus view $x'$ on a variable $x$ behaves as if
executed on $-x$: $\xgetmax{x'}$ is defined as $-\xgetmin{x}$ and
$\xadjmax{x'}{n}$ is defined as $\xadjmin{x}{-n}$. With views, the
implementation of the maximum propagator can be reused: we \emph{derive} a
propagator for the minimum constraint by instantiating the maximum propagator
with minus views for its variables.

The feasibility of variable views rests on today's programming
languages' support for generic (or polymorphic) constructions (for
example, templates in \CPP) and that the simple transformations
provided by views are optimized away. 

\pparagraph{Contributions}
This paper contributes an implementation independent model for
views and derived propagators, techniques for deriving
propagators, and an evaluation that shows that views are widely
applicable, drastically reduce programming effort, and are more
efficient than decomposition.

More specifically, the key contribution is the identification of properties of
views that are essential for deriving \emph{perfect} propagators. To this end,
the paper establishes a formal model that defines a view as a function and a
derived propagator as functional composition of views (mapping values to
values) with a propagator (mapping variable domains to variable domains). This
model yields all the desired results: derived propagators are indeed
propagators; derived propagators faithfully implement the intended
constraints; domain consistency carries over to derived propagators; different
forms of bounds consistency over integer variables carry over provided that
the views satisfy additional properties.

After establishing the fundamental results, we address further properties of
derived propagators such as idempotence, subsumption, and events. Finally, we
clarify the connection between derived propagators and path consistency when
regarding views as binary constraints.

We introduce techniques for deriving propagators that use views for
specialization and generalization of propagators, channeling between variable
domains, and general domain-specific transformations. We show how to apply
these techniques for different variable domains using various views. We
provide a breakdown of how successful the use of derived propagators has been
for Gecode.

\pparagraph{Overview}
The next section introduces the basic notions we will use. Sect.~\ref{sec:views_and_derived_propagators} presents views and derived propagators and proves fundamental properties like correctness and completeness. The following three sections develop techniques for deriving propagators: transformation, generalization, specialization, and channeling. Sect.~\ref{sec:extended_properties_of_derived_propagators} presents extensions of the model, and Sect.~\ref{sec:limitations} discusses its limitations. Sect.~\ref{sec:experiments} provides empirical evidence that views are useful in practice.

\section{Preliminaries}
\label{sec:preliminaries}

This section sets the stage for the paper with definitions of the basic concepts.

\pparagraph{Variables and constraints}
We assume a finite set of variables $\Var=\{x_1,\dots,x_n\}$ and a finite set
of values $\Val$. Constraints are characterized by assignments $a\in\Asn$
that map variables to values: $\Asn=\fun{\Var}{\Val}$. A constraint $c\in\Con$
is a relation over the variables, represented as the set of all assignments that satisfy the constraint, $\Con=\Power{\Asn}$. We base constraints on full assignments, defined for all variables in $\Var$. However, for typical constraints, only a subset $\vars(c)$ of the variables is \emph{significant}; the constraint is the full relation for all $x\notin\vars(c)$. We write a constraint in extension ($c=\{(x\mapsto 0,y\mapsto 1),(x\mapsto 1,y\mapsto 2)\}$) or intensionally ($c\equiv x<y$).

\pparagraph{Domains}
Constraints are implemented by propagators over domains, which are constructed as follows. A \emph{domain} $d\in\Dom$ maps each variable to a finite set of possible values, the \emph{variable domain} $d(x)\subseteq\Val$.

A domain $d$ can be identified with a set of assignments $d\in\Power{\Asn}$. We can therefore treat domains as constraints. In particular, for any assignment $a$, $\{a\}$ is a domain as well as a constraint. We simply write \emph{domain} for domains and variable domains when there is no risk of confusion.

A domain $d_1$ is \emph{stronger} than a domain $d_2$ (written $d_1\stronger d_2$), iff for all variables $x$, $d_1(x)\subseteq d_2(x)$. By $\dom{c}$ we refer to the strongest domain including all valid assignments of a constraint, defined as $\min\setc{d\in\Dom}{c\subseteq d}=\setc{a}{\forall x\ \exists b\in c.\ a(x)=b(x)}$. The minimum exists as domains are closed under intersection, and the definition is non-trivial  because not every constraint can be captured by a domain. Now, for a constraint $c$ and a domain $d$, $\dom{c\cap d}$ refers to removing all values from $d$ not supported by the constraint~$c$.

\pparagraph{Propagators}
Propagators serve here as implementations of constraints.  They are sometimes also referred to as constraint narrowing operators or filter functions. A propagator is a function $p\in\fun{\Dom}{\Dom}$ that is contracting ($p(d)\stronger d$) and monotone ($d'\stronger d\mimpl p(d')\stronger p(d)$).  Idempotence is not required.

Propagators are contracting, they only remove values from variable domains. For an assignment $a$, a propagator $p$ hence has only two options: accept it ($p(\{a\})=\{a\}$), or reject it ($p(\{a\})=\emptyset$). Monotonicity guarantees that if some domain $d$ contains an assignment $a\in d$ that $p$ accepts, then $p$ will not remove $a$ from $d$: $a\in p(d)$. The propagator therefore behaves like a characteristic function for the set of accepted assignments. This set is the \emph{associated constraint of $p$}.

We say that a propagator $p$ \emph{implements its associated constraint} $c_p=\setc{a\in\Asn}{p(\{a\})=\{a\}}$. Monotonicity implies that for any domain $d$, we have $\dom{c_p\cap d}\stronger p(d)$: no solution of $c_p$ is ever removed by $p$. We say that $p$ is \emph{sound} for any $c\subseteq c_p$ and \emph{weakly complete} for any $c'\supseteq c_p$ (meaning that it accepts all assignments in $c$ and rejects all assignments not in $c'$). For any constraint $c$, we can find at least one propagator $p$ such that $c=c_p$. Typically, there are several propagators, differing by \emph{propagation strength} (see Sect.~\ref{sec:completeness}).

Our definitions of soundness and different notions of completeness for propagators are based on and equivalent to Benhamou's~\cite{BenhamouHeterogeneous} and Maher's~\cite{Maher:ICLP:02}.
We specify \emph{what} is computed by constraint propagation and not \emph{how}. Approaches for performing constraint propagation can be found in~\cite{BenhamouHeterogeneous,AptPrinciples:2003,SchulteStuckey:TOPLAS:2007}.

\section{Views and Derived Propagators}
\label{sec:views_and_derived_propagators}

We now introduce our central concepts, views and derived propagators.

A \emph{view} on a variable $x$ is an injective function $\varphi_x\in\fun{\Val}{\Val'}$, mapping values from $\Val$ to values from a possibly different set $\Val'$. We lift a family of views $\varphi_x$ (one for each $x\in\Var$) point-wise to assignments as follows: $\varphi_\Asn(a)(x)=\varphi_x(a(x))$. Finally, given a family of views lifted to assignments, we define a view $\varphi\in\fun{\Con}{\Con}$ on constraints as $\varphi(c)=\setc{\varphi_\Asn(a)}{a\in c}$. The inverse of that view is defined as $\varphi^-(c)=\setc{a\in\Asn}{\varphi_\Asn(a)\in c}$.

In the implementation, a view on $x$ presents the same interface as $x$, but applies transformations when a propagator adjusts or accesses the domain of $x$ through the view. In our model, $\varphi$ performs the transformations for accessing, and $\varphi^-$ for adjusting the variable domains.
Views can now be composed with a propagator: a \emph{derived propagator} is defined as $\varphihat(p)(d)=\varphi^-(p(\varphi(d)))$, or, using function composition, as $\varphihat(p)=\varphi^-\circ p\circ\varphi$.

\begin{example}
Given a propagator $p$ for the constraint $c\equiv(x=y)$, we want to derive a propagator for $c'\equiv(x=2y)$ using a view $\varphi$ such that $\varphi^-(c)=c'$.

It is usually easier to think about the other direction: $\varphi(c')\subseteq c$. Intuitively, the function $\varphi$ leaves $x$ as it is and scales $y$ by 2, while $\varphi^-$ does the inverse transformation. We thus define $\varphi_x(v)=v$ and $\varphi_y(v)=2v$.
We have a subset relation because some tuples of $c$ may be ruled out by $\varphi$. For instance, with $\varphi$ defined as above, there is no assignment $a$ such that $\varphi_\Asn(a)(y) = 3$, but the assignment $(x\mapsto 3,y\mapsto 3)$ is in $c$.

This example also makes clear why the set $\Val'$ is allowed to differ from $\Val$. In this particular case, $\Val'$ has to contain all multiples of $2$ of elements in $\Val$.

The derived propagator is $\varphihat(p)=\varphi^-\circ p\circ\varphi$. 
We say that $\varphihat(p)$ ``uses a scale view on'' $y$, meaning that
$\varphi_y$ is the function defined as
$\varphi_y(v)=2v$. Similarly, using an identity view on $x$
amounts to $\varphi_x$ being the identity function on $\Val$.

Given the assignment $a=(x\mapsto 2,y\mapsto 1)$, we first apply $\varphi_\Asn$ and get $\varphi_\Asn(a)=(x\mapsto 2,y\mapsto 2)$. This is accepted by $p$ and returned unchanged, so $\varphi^-$ transforms it back to $a$. Another assignment, $a'=(x\mapsto 1,y\mapsto 2)$, is transformed to $\varphi_\Asn(a')=(x\mapsto 1,y\mapsto 4)$, rejected ($p(\{\varphi_\Asn(a')\})=\emptyset$), and the empty domain is mapped to the empty domain by $\varphi^-$. The propagator $\varphihat(p)$ implements $\varphi^-(c)$.
\end{example}

Views and derived propagators satisfy a number of essential properties:
\begin{enumerate}
  \item A derived propagator $\varphihat(p)$ is in fact a propagator.
  \item The associated constraint of $\varphihat(p)$ is $\varphi^-(c_p)$.
  \item A view $\varphi$ preserves contraction of a propagator $p$: If $p(\varphi(d))\subset \varphi(d)$, then $\varphihat(p)(d)\subset d$. This property makes sure that if the propagator makes an inference, then this inference will actually be reflected in a domain change.
\end{enumerate}
In the following, we will prove these properties. For the proofs, we employ some direct consequences of the definitions of views and derived propagators:
(1) $\varphi$ and $\varphi^-$ are monotone by construction; (2) $\varphi^-\circ\varphi=\mathrm{id}$ (the identity function); (3) $|\varphi(\{a\})|=1$, $\varphi(\emptyset)=\emptyset$; (4) for any view $\varphi$ and domain $d$, we have $\varphi(d)\in\Dom$ and $\varphi^-(d)\in\Dom$ (as views are defined point-wise).

\begin{theorem}
\label{theorem:derivedprop}
A derived propagator is a propagator: for all propagators $p$ and views $\varphi$, $\varphihat(p)$ is a monotone and  contracting function in $\fun{\Dom}{\Dom}$.
\end{theorem}

\begin{proof}
The derived propagator is well-defined because both $\varphi(d)$ and $\varphi^-(d)$ are domains (see (4) above). Monotonicity is obvious, as compositions of monotone functions are monotone.
For contraction, we have $p(\varphi(d))\stronger\varphi(d)$ as $p$ is contracting. By monotonicity of $\varphi^-$, we know that $\varphi^-(p(\varphi(d)))\stronger\varphi^-(\varphi(d))$. As $\varphi^-\circ\varphi=\mathrm{id}$, we have $\varphi^-(p(\varphi(d)))\stronger d$, which proves that $\varphihat(p)$ is contracting.
In summary, for any propagator $p$, $\varphihat(p)=\varphi^-\circ p\circ\varphi$ is a propagator.  
\end{proof}

\begin{theorem}
If $p$ implements $c_p$, then $\varphihat(p)$ implements $\varphi^-(c_p)$.
\end{theorem}

\begin{proof}
As $p$ implements $c_p$, we know $p(\{a\})= c_p\cap\{a\}$ for all assignments $a$. With $|\varphi(\{a\})|=1$, we have $p(\varphi(\{a\}))= c_p\cap\varphi(\{a\})$.  Furthermore, we know that $c_p\cap\varphi(\{a\})$ is either $\emptyset$ or $\varphi(\{a\})$. Case $\emptyset$: We have $\varphi^-(p(\varphi(\{a\})))=\emptyset=\{a\}\cap\varphi^-(c_p)$. Case $\varphi(\{a\})$: As $\varphi^-\circ\varphi=\mathrm{id}$, we have $\varphi^-(p(\varphi(\{a\})))=\{a\}$. Furthermore:
    $$\begin{array}{clcl}
    & c_p\cap\varphi(\{a\})=\varphi(\{a\}) & \quad\mimpl\quad
    & \exists b\in c_p.\ b=\varphi(a)\\
    \mimpl & a \in \setc{a'\in\Asn}{\varphi(a')\in c_p}
    & \mimpl & a\in\varphi^-(c_p)
    \end{array}$$
Together, this shows that $\varphi^-\circ p\circ\varphi(\{a\})=\{a\}\cap\varphi^-(c_p)$.
\end{proof}

\begin{theorem}
\label{theorem:contraction}
Views preserve contraction: for any domain $d$, if $p(\varphi(d))\stronger \varphi(d)$, then $\varphihat(p)(d)\strstronger d$.
\end{theorem}

\begin{proof}
Recall the definition of $\varphi^-(c)$ as $\setc{a\in\Asn}{\varphi_\Asn(a)\in c}$. It clearly follows that $|\varphi^-(c)|\leq|c|$. Similarly, we know that $|\varphi(c)|=|c|$. From $p(\varphi(d))\subset \varphi(d)$, we know that $|p(\varphi(d))|<|\varphi(d)|$. Together, this yields $|\varphihat(p)(d)|<|\varphi(d)|=|d|$. We have already seen in Theorem~\ref{theorem:derivedprop} that $\varphihat(p)(d)\stronger d$, so we can conclude that $\varphihat(p)(d)\strstronger d$.
\end{proof}

\pparagraph{Completeness}
\label{sec:completeness}
Weak completeness, as introduced above, is the minimum required for a constraint solver to be complete. A weakly complete propagator does not have to prune variable domains, it only has to check if an assigned domain is a solution of the constraint. The success of constraint propagation however crucially depends on strong propagators that prune variable domains.

The strongest possible inference that a single propagator can do establishes \emph{domain consistency} (also known as \emph{generalized arc consistency}): a domain $d$ is domain consistent for a constraint $c$, iff for all variables $x_i$ and all values $v_i\in d(x_i)$, there exist values $v_j\in d(x_j)$ for all other variables $x_j$ such that the assignment $(x_1\mapsto v_1,\dots,x_i\mapsto v_i,\dots,x_n\mapsto v_n)$ is a solution of $c$.

A propagator is \emph{domain complete} (or simply complete) for a constraint $c$ if it establishes domain consistency. More formally, a propagator $p$ is complete for a constraint $c$ iff for all domains $d$, we have $p(d)\stronger\dom{c\cap d}$. A complete propagator thus removes all assignments from $d$ that are inconsistent with $c$.

We will now prove that propagators derived from complete propagators are also  complete. In Sect.~\ref{sec:views_for_integer_variables}, we will extend this result to weaker notions of completeness, such as \boundsz{} and \boundsr{} completeness.

For this proof, we need two auxiliary definitions. A constraint $c$ is a \emph{$\varphi$ constraint} iff for all $a\in c$, there is a $b\in\Asn$ such that $a=\varphi_\Asn(b)$. A view $\varphi$ is \emph{\dominj} iff $\varphi^-(\dom{c})=\dom{\varphi^-(c)}$ for all $\varphi$ constraints $c$.

For the completeness proof, we need a lemma that states that any view is \dominj.

\begin{proof}
By definition of $\varphi^-$ and $\dom{\cdot}$, we have
$\varphi^-(\dom{c})=\setc{a\in\Asn}{\forall x.\exists b\in
c.\varphi_\Asn(a)(x)=b(x)}$. As $c$ is a $\varphi$ constraint, we can find
such a $b$ that is in the range of $\varphi_\Asn$, if and only if there is
also a $b'\in\varphi^-(c)$ such that $\varphi_\Asn(b')=b$. Therefore, we get
$\setc{a\in\Asn}{\forall x.\exists
b'\in\varphi^-(c).a(x)=b'(x)}=\dom{\varphi^-(c)}$.
\end{proof}

Furthermore, we need a lemma that states that views commute with set intersection: For any view $\varphi$, the equation $\varphi^-(c_1\cap c_2)=\varphi^-(c_1)\cap\varphi^-(c_2)$ holds.

\begin{proof}
By definition of $\varphi^-$, we have $\varphi^-(c_1\cap c_2)=\setc{a\in\Asn}{\varphi_\Asn(a)\in c_1\land \varphi_\Asn(a)\in c_2}$. As $\varphi_\Asn$ is a function, this is equal to $\setc{a\in\Asn}{\varphi_\Asn(a)\in c_1}\cap\setc{a\in\Asn}{\varphi_\Asn(a)\in c_2}=\varphi^-(c_1)\cap\varphi^-(c_2)$.
\end{proof}

\begin{theorem}
  \label{thm:completeness}
If $p$ is complete for $c$, then $\varphihat(p)$ is complete for $\varphi^-(c)$.
\end{theorem}

\begin{proof}
By monotonicity of $\varphi$ and completeness of $p$, we know that $\varphi^-\circ p\circ\varphi(d) \stronger \varphi^-(\dom{c \cap \varphi(d)})$. We now use the fact that $\varphi^-$ is \dominj{} and commutes with set intersection:
  \begin{eqnarray*}
  &\varphi^-(\dom{c \cap \varphi(d)}) = \dom{\varphi^-(c \cap \varphi(d))} =\\
  &\dom{\varphi^-(c) \cap \varphi^-(\varphi(d))} = \dom{\varphi^-(c) \cap d}
  \end{eqnarray*}

\end{proof}

\section{Boolean Variables: Transformation}
\label{sec:views_for_boolean_variables}

This section discusses views and derived propagators for Boolean
variables where $\Val=\{0,1\}$. Not surprisingly, the only view apart
from identity for Boolean variables captures negation. That is, using a
\emph{negation view} on $x$ defines $\varphi_x(v)=1-v$ for $x\in\Var$
and $v\in\Val$.

Negation views are more widely applicable than one would initially
believe. They demonstrate how views can be used systematically to obtain implementations of constraint variants by \emph{transformation}.

\pparagraph{Boolean connectives}
The immediate application of negation views is to derive propagators
for all Boolean connectives from just three propagators: A
negation view for $x$ in $x=y$ yields a propagator
for $\neg x=y$. From disjunction $x\vee y=z$ one can derive
conjunction $x\wedge y=z$ with negation views on $x$, $y$, $z$, and
implication $x\rightarrow y=z$ with a negation view on $x$. From
equivalence $x\leftrightarrow y=z$ one can derive exclusive or $x\xor
y=z$ with a negation view on $z$.

As Boolean constraints are widespread in models, it pays off to
optimize frequently occurring cases. One important propagator is
disjunction $\bigvee_{i=1}^n x_i=y$ for arbitrarily many
variables; again conjunction can be derived with negation views
on the $x_i$ and on $y$. Another important propagator is for the
constraint $\bigvee_{i=1}^n x_i=1$, stating that the disjunction must
be true. A propagator for this constraint is essential as the
constraint occurs frequently and as it can be implemented efficiently
using watched literals, see for example~\cite{minion:wl}. With views
and derived propagators all implementation work is readily reused
for conjunction. This shows a general advantage of views: effort put into optimizing a single propagator directly pays off for all other propagators derived from it.

\pparagraph{Boolean cardinality}
Like the constraint $\bigvee_{i=1}^n x_i=1$, the Boolean cardinality
constraint $\sum_{i=1}^n x_i \geq c$ occurs frequently and can be
implemented efficiently using watched literals (requiring $c+1$
watched literals, Boolean disjunction corresponds to the case where
$c=1$). But also a propagator for $\sum_{i=1}^n x_i \leq c$ can be
derived using negation views with the following transformation: 
$$
\begin{array}{rclcl}
\sum_{i=1}^n x_i \leq c &\iff& -\sum_{i=1}^n x_i \geq -c 
&\iff& n-\sum_{i=1}^n x_i \geq n-c \\
&\iff& \sum_{i=1}^n 1-x_i \geq n-c 
&\iff& \sum_{i=1}^n \neg x_i \geq n-c \\
\end{array}
$$

\pparagraph{Reification}
Many reified constraints (such as $\reify{\sum_{i=1}^n x_i=c}{b}$)
also exist in a negated version (such as $\reify{\sum_{i=1}^n x_i\neq
  c}{b}$). Deriving the negated version is trivial by using a negation
view on the Boolean control variable $b$. This contrasts nicely with
the effort without views: either the entire code must be duplicated or
the parts that perform checking whether the constraint or its negation
is entailed must be factorized out and combined differently for the
two variants.

\section{Integer Variables: Generalization, Bounds Consistency, Specialization}
\label{sec:views_for_integer_variables}

Common views for finite domain integer variables capture linear
transformations of the integer
values. In~\cite{SchulteTack:Advances:2006}, the following views are
introduced for a variable $x$ and values $v$: a \emph{minus view} on
$x$ is defined as $\varphi_x(v)=-v$, an \emph{offset view} for
$o\in\ZZ$ on $x$ is defined as $\varphi_x(v)=v+o$, and a
\emph{scale view} for $a\in\ZZ$ on $x$ is defined as
$\varphi_x(v)=a\cdot v$.

Propagators for integer variables offer a greater degree of freedom
concerning their level of completeness. While Boolean propagators most
often will be domain complete, bounds completeness is important for
integer propagators. Before we discuss transformation and
generalization techniques for deriving integer propagators, we study
how bounds completeness is affected by views.

\pparagraph{Bounds consistency and bounds completeness}
There are several different notions of bounds consistency in
the literature (see~\cite{ChoiHarveyLeeStuckey:AI:2006} for an
overview). For our purposes, we distinguish \boundsd, \boundsz,
and \boundsr{} consistency:
\begin{itemize}
\item A domain $d$ is \emph{\boundsd{} consistent} for a constraint
  $c$, iff for all variables $x_i$ there exist $v_j\in d(x_j)$ for all other 
  variables $x_j$ such that
  $\{x_1\mapsto v_1,\dots,x_i\mapsto\min(d(x_i)),\dots,x_n\mapsto
  v_n\}\in c$ and analogously for $x_i\mapsto\max(d(x_i))$.
\item A domain $d$ is \emph{\boundsz{} consistent} for a constraint
  $c$, iff for all variables $x_i$, there exist integers $v_j$ with
  $\min(d(x_j))\leq v_j\leq\max(d(x_j))$ for all other variables $x_j$
  such that $\{x_1\mapsto
  v_1,\dots,x_i\mapsto\min(d(x_i)),\dots,x_n\mapsto v_n\}\in c$ and
  analogously for $x_i\mapsto\max(d(x_i))$.
\item A domain $d$ is \emph{\boundsr{} consistent} for a constraint
  $c$, iff for all variables $x_i$, there exist real numbers
  $v_j\in\RR$ with $\min(d(x_j))\leq v_j\leq\max(d(x_j))$ for all
  other variables $x_j$ such that $\{x_1\mapsto
  v_1,\dots,x_i\mapsto\min(d(x_i)),\dots,x_n\mapsto v_n\}\in c_\RR$
  and analogously for $x_i\mapsto\max(d(x_i))$, where $c_\RR$ is $c$
  relaxed to $\RR$ (for constraints like arithmetics where relaxation makes sense).
\end{itemize}
A propagator $p$ is bounds($X$) complete for its associated constraint
$c_p$, iff $p(d)$ is bounds($X$) consistent for $c_p$ for every domain
$d$ that is a fixpoint of $p$.
We use an equivalent definition based on the \emph{strongest convex domain} that contains a constraint, $\conv(c)=\min\setc{d\in\Dom}{c\stronger d\text{ and }d\text{ convex}}$. A convex domain maps each variable to an interval, so that
$\conv(c)(x)=\{\min_{a\in c}(a(x)),\dots,\max_{a\in c}(a(x))\}$.
Note that $\conv(c)$ is weaker than the strongest domain that contains $c$: $\conv(c)\supseteq\dom{c}$ for all constraints $c$. In the same way as Benhamou~\cite{BenhamouHeterogeneous} and Maher~\cite{Maher:ICLP:02}, we define
\begin{itemize}
\item $p$ is \boundsd{} complete for $c$ iff
  $p(d)\stronger\conv(c\cap d)$.
\item $p$ is \boundsz{} complete for $c$ iff
  $p(d)\stronger\conv(c\cap \conv(d))$.
\item $p$ is \boundsr{} complete for $c$ iff
  $p(d)\stronger\conv(c_\RR\cap \conv_\RR(d))$, where
  $\conv_\RR(d)$ is the convex hull of $d$ in $\RR$, and
  $c_\RR$ is $c$ relaxed to $\RR$.
\end{itemize}

\pparagraph{Bounds completeness of derived propagators}
Theorem~\ref{thm:completeness} states that propagators derived from
domain complete propagators are domain complete.  A similar theorem 
holds for bounds completeness, if views commute with $\conv(\cdot)$ in the following ways:

A view $\varphi$ is \emph{interval injective} iff $\varphi^-(\conv(c))=\conv(\varphi^-(c))$ for all $\varphi$ constraints $c$. It is \emph{interval bijective} iff it is interval injective and $\varphi(\conv(d))=\conv(\varphi(d))$ for all domains $d$.

Proving bounds completeness of derived propagators is now similar to proving 
domain completeness.  We only formulate \boundsz{} completeness.

\begin{theorem}
  If $p$ is \boundsz{} complete for $c$ and $\varphi$ is interval
  bijective, then $\varphihat(p)$ is \boundsz{} complete for
  $\varphi(c)$.
\end{theorem}

\begin{proof}
By monotonicity of $\varphi$ and \boundsz{} completeness of $p$, we
know that $\varphi^-\circ p\circ\varphi(d) \stronger
\varphi^-(\conv(c \cap \conv(\varphi(d))))$. We now use the fact that both $\varphi$ and $\varphi^-$ commute with $\conv$ and intersection:
\begin{eqnarray*}
&\varphi(\conv(c \cap \conv(\varphi^{-1}(d)))) = \varphi(\conv(c \cap \varphi^{-1}(\conv(d)))) =\\
&\conv(\varphi(c \cap \varphi^{-1}(\conv(d)))) = \conv(\varphi(c) \cap \varphi(\varphi^{-1}(\conv(d)))) = \\
&\conv(\varphi(c) \cap \conv(d))
\end{eqnarray*}

\end{proof}

The proof for \boundsd{} is analogous, but we only require interval
injectivity for the view. With an interval injective view, one can also derive
\boundsr{} complete propagators from \boundsr{} or \boundsz{} complete
propagators. Table~\ref{tab:completenessviewbijectivity} summarizes how
completeness depends on view bijectivity.

\begin{table}
\caption{Completeness of derived propagators}
\label{tab:completenessviewbijectivity} 
  \footnotesize
  \centering
        \begin{tabular}{|l||c|c|c|}
        \hline
        \emph{propagator}\quad & \multicolumn{3}{c|}{\emph{view}}\\
        \cline{2-4}
                 & \emph{interval bijective} & \emph{interval injective} & \emph{arbitrary}\\
        \hline\hline
        domain    & domain    & domain   & domain\\
        \boundsd  & \boundsd  & \boundsd & weakly\\
        \boundsz  & \boundsz  & \boundsr & weakly\\
        \boundsr  & \boundsr  & \boundsr & weakly\\
        \hline
        \end{tabular}
\end{table}

The views for integer variables presented at the beginning of this
section have the following properties: minus and offset views are interval bijective, whereas a scale view for $a\in\ZZ$ on $x$
is always interval injective and only interval bijective if $a=1$ or $a=-1$ (in which cases it coincides with the identity view or a minus view, respectively). An important consequence is that a \boundsz{} complete propagator for the constraint $\sum_{i}x_i=c$, when instantiated with scale views for the $x_i$, results in a \boundsr{} complete propagator for $\sum_i a_ix_i=c$.

\pparagraph{Transformation}
Like the negation view for Boolean variables, minus views for integer
variables help to derive propagators following simple
transformations: for example, $\min(x,y)=z$ can be derived from
$\max(x,y)=z$ by using minus views for $x$, $y$, and
$z$.

Transformations through minus views can improve performance in subtle
ways. Consider a $\boundsz$ consistent propagator for multiplication
$x\times y=z$. Propagation depends on whether zero is still included
in the domains of $x$, $y$, or $z$. Testing for inclusion of zero each
time the propagator is executed is not very efficient. Instead, one
would like to rewrite the propagator to special variants where $x$,
$y$, and $z$ are either strictly positive or negative. These variants
can propagate more efficiently, in particular because propagation can
easily be implemented to be idempotent (see
Section~\ref{sec:extended_properties_of_derived_propagators}). Implementing
three different propagators (all variables strictly positive, $x$ or
$y$ strictly positive, only $z$ strictly positive) seems
excessive. Here, a single propagator assuming that all views are
positive is sufficient, the others can be derived using minus views.

\pparagraph{Generalization}
Offset and scale views are useful for generalizing
propagators. Generalization has two key advantages: simplicity and
efficiency. A more specialized propagator is often simpler to implement than
a generalized version. The possibility to use the specialized version
when the full power of the general version is not required may save
space and time during execution.

The propagator for a linear equality constraint
$\sum_{i=1}^n x_i=c$ is efficient for the common case that the linear
equation has only unit coefficients. The more general case
$\sum_{i=1}^n a_i x_i=c$ can be derived by using scale views for $a_i$
on $x_i$ (This of course also holds true for linear
inequality and disequality rather than equality). Similarly, a propagator for $\distinct{x_i}$ can be
generalized to $\distinct{c_i+x_i}$ by using offset views for
$c_i\in\ZZ$ on $x_i$. Likewise, a propagator for the element
constraint $\element{\langle c_1,\ldots,c_n\rangle}{x}=y$ can be
generalized to $\element{\langle c_1,\ldots,c_n\rangle}{x+o}=y$ with
an offset view, where $o\in\ZZ$ provides a useful offset for the index
variable $x$.
It is important to recall that propagators are derived: in Gecode, the above generalizations are applied to domain as well as
bounds complete propagators.

\pparagraph{Specialization}
We employ \emph{constant views} to specialize propagators. A constant view behaves like a fixed variable. In practice, specialization has two advantages: Fewer variables are needed, which means less space consumption. And specialized propagators can be compiled to more efficient code, if constants are known at compile time.

Examples for specialization are a propagator for binary linear inequality $x+y\leq c$ derived from a propagator for $x+y+z\leq c$ by using a constant 0 for $z$; a Boolean propagator for $x\land y\lequiv 1$ from $x\land y\lequiv z$ and constant 1 for $z$; a propagator for the element constraint $\element{\langle c_1,\ldots,c_n\rangle}{y}=z$ derived from a propagator for $\element{\langle x_1,\ldots,x_n\rangle}{y}=z$; a reified propagator for $(x=c)\lequiv b$ from $(x=y)\lequiv b$ and a constant $c$ for $y$; a propagator for counting $|\setc{i}{x_i=y}|=c$ from a propagator for $|\setc{i}{x_i=y}|=z$; and many more.

We have to extend our model to support constant views. Propagators may now be defined with respect to a superset of the variables, $\Var'\supseteq\Var$. A constant view for the value $k$ on a variable $z\in\Var'\setminus\Var$ translates between the two sets of variables as follows:
$$
\begin{array}{rcl}
  \varphi^-(c) &=& \setc{\restrict{a}{\Var}}{a\in c}\\
  \varphi(c) &=& \setc{a[k/z]}{a\in c}\\
\end{array}
$$
Here, $a[k/z]$ means augmenting the assignment $a$ so that it maps $z$ to $k$, and $\restrict{a}{\Var}$ is the functional restriction of $a$ to the set $\Var$.
It is important to see that this definition preserves failure: if a propagator returns a failed domain $d$ that maps $z$ to the empty set, then $\varphi^-(d)$ is the empty set, too.

\pparagraph{Indexicals}
Views that perform arithmetic transformations are related to indexicals~\cite{CarlssonOttossonEa:97,VanHentenryckSaraswatEa:98}. An indexical is a propagator that prunes a single variable and is defined in terms of range expressions. A view is similar to an indexical with a single input variable. However, views are not used to build propagators directly, but to derive new propagators from existing ones. Allowing the full expressivity of indexicals for views would imply giving up our completeness results.

Another related concept are arithmetic expressions, which can be used for modeling in many systems (such as ILOG Solver~\cite{PugetLeconte:95}). In contrast to views, these expressions are not used for propagation directly and, like indexicals, yield no completeness guarantees.

\section{Set Variables: Channeling}
\label{sec:views_for_set_variables}

Set constraints deal with variables whose domains are sets of finite sets. This powerset lattice is a Boolean algebra, so typical constraints are constructed from the Boolean primitives disjunction (union), conjunction (intersection), and negation (complement), and the relations equality and implication (subset).

\pparagraph{Transformation and Specialization}
As for Boolean and integer variables, views on set variables enable transformation and specialization. Using \emph{complement views} (analogous to Boolean negation) on $x,y,z$ with a propagator for $x\cap y=z$ yields a propagator for  $x\cup y=z$. A complement view on $y$ gives us $x\setminus y=z$. Constant views like the empty set or the universe enable specialization; for example, $x \cap y = z$ implements set disjointness if
$z$ is the constant empty set.

\pparagraph{Channeling views}
A channeling view changes the type of the values that a variable can take. Our model already accommodates for this as a view $\varphi_x$ maps elements between different sets $\Val$ and $\Val'$.

An important channeling view is a \emph{singleton view} on an integer variable $x$, defined as $\varphi_x(v)=\{v\}$. It presents an integer variable as a singleton set variable. Many useful constraints involve both integer and set variables, and some of them can be expressed with singleton views. The simplest constraint is $x\in y$, where $x$ is an integer variable and $y$ a set variable. Singleton views let us implement it as $\{x\}\subseteq y$, and just as easily give us the negated and reified variants. Obviously, this extends to $\{x\}\diamond y$ for all other set relations $\diamond$.

Singleton views can also be used to derive pure integer constraints from
set propagators. For example, the constraint
$\text{same}([x_1,\dots,x_n],[y_1,\dots,y_m])$ states that the two sequences
of integer variables take the same values. With singleton views,
$\bigcup_{i=1}^n \{x_i\} = \bigcup_{j=1}^m \{y_j\}$ implements this 
constraint.

\pparagraph{Channeling between domain implementations}
Most systems approximate finite set domains as convex sets defined by a lower
and an upper bound~\cite{Gervet:97}. However, Hawkins et
al.~\cite{Hawkins:JAIR:2005} introduced a complete representation for the
domains of finite set variables using ROBDDs. Channeling views can translate
between interval- and ROBDD-based implementations. We can derive a propagator
on ROBDD-based variables from a set-interval propagator, and thus reuse
set-interval propagators for which no efficient ROBDD representation exists.

\section{Extended Properties of Derived Propagators}
\label{sec:extended_properties_of_derived_propagators}

This section discusses how views can be composed, how derived propagators behave with respect to idempotence and subsumption, and how events can be used to schedule derived propagators. Finally, we discuss the relation between views and path consistency.

\pparagraph{Composing views}
A derived propagator permits further derivation: $\varphihat(\varphihatprime(p))$ for two views $\varphi,\varphi'$ is perfectly acceptable, properties like correctness and completeness carry over. For instance, we can derive a propagator for $x-y=c$ from a propagator for $x+y=0$ by combining an offset 
view and a minus view on $y$.

\pparagraph{Idempotent propagators}
A propagator is idempotent iff $p(p(d)) = p(d)$ for all domains $d$. Some systems require all propagators to be idempotent, others apply optimizations if the idempotence of a propagator is known~\cite{SchulteStuckey:TOPLAS:2007}.
If a propagator is derived from an idempotent propagator, the result is idempotent again:

\begin{theorem}
If $p(p(d))=p(d)$ for a propagator $p$ and a domain $d$, then, for any view $\varphi$, $\varphihat(p)(\varphihat(p)(d))=\varphihat(p)(d)$.
\end{theorem}

\begin{proof}
Function composition is associative, so we can write $\varphihat(p)(\varphihat(p)(d))$ as $\varphi^-\circ p\circ(\varphi\circ\varphi^-)\circ p\circ\varphi(d)$. We know that $\varphi\circ\varphi^-=\mathrm{id}$ for all domains that contain only assignments on which $\varphi^-$ is fully defined, meaning that $|\varphi^-(d)|=|d|$. As we first apply $\varphi$, this is the case here, so we can remove $\varphi\circ\varphi^-$, leaving $\varphi^-\circ p\circ p\circ\varphi(d)$. As $p$ is idempotent, this is equivalent to $\varphi^-\circ p\circ\varphi(d)=\varphihat(p)(d)$.
\end{proof}

\pparagraph{Subsumption}
A propagator is subsumed for a domain $d$ iff for all stronger domains $d'\subseteq d$, $p(d')=d'$. Subsumed propagators do not contribute any propagation in the remaining subtree of the search, and can therefore be removed. Deciding subsumption is coNP-complete in general, but for most propagators an approximation can be decided easily. This can be used to optimize propagation.

\begin{theorem}
  \label{theorem:subsumtion}
$p$ is subsumed by $\varphi(d)$ iff $\varphihat(p)$ is subsumed by $d$.
\end{theorem}

\begin{proof}
The definition of $\varphi$ gives us that $\forall d'\subseteq d.\ \varphi^-(p(\varphi(d'))) = d'$ is equivalent with
$\forall d'\subseteq d.\ \varphi^-(p(\varphi(d'))) = \varphi^-(\varphi(d')$. As $\varphi^-$ is a function, and because it is contraction-preserving (see Theorem~\ref{theorem:contraction}), this is equivalent with
$\forall d'\subseteq d.\ p(\varphi(d')) = \varphi(d')$.
Because all $\varphi(d')$ are subsets of $\varphi(d)$, we can rewrite this to
$\forall d''\subseteq \varphi(d).\ p(d'') = d''$, concluding the proof.
\end{proof}

\pparagraph{Events}
Many systems control propagator invocation using \emph{events} (for a detailed discussion, see~\cite{SchulteStuckey:TOPLAS:2007}).  An event describes how a domain changed.  Typical events for finite domain integer variables are:
the variable $x$ becomes fixed ($\mathrm{fix}(x)$);
the lower bound of variable $x$ changes ($\mathrm{lbc}(x)$);
the upper bound of variable $x$ changes ($\mathrm{ubc}(x)$);
the domain of variable $x$ changes ($\mathrm{dmc}(x)$).
In some systems, $\mathrm{lbc}(x)$ and $\mathrm{ubc}(x)$ are collapsed into one event, $\mathrm{bc}(x)=\mathrm{lbc}(x)\lor\mathrm{ubc}(x)$.
Events are monotone: if $\mathrm{events}(d,d'')$ is the set of events occurring when the domain changes from $d$ to $d''$ (with $d''\stronger d$), then we have $\mathrm{events}(d,d'')=\mathrm{events}(d,d')\cup\mathrm{events}(d',d'')$ for any $d''\stronger d'\stronger d$.
Propagators are associated with \emph{event sets}: A propagator $p$ depends on an event set $\mathit{es}(p)$ iff
\begin{enumerate}
        \item for all $d$ if $p(d)\neq p(p(d))$, then $\mathrm{events}(d,p(d))\cap\mathit{es}(p)\neq\emptyset$
        \item for all $d,d'$ where $p(d)=d$, $d'\stronger d$, $p(d')\neq d'$, then $\mathrm{events}(d,d')\cap\mathit{es}(p)\neq\emptyset$
\end{enumerate}
If a propagator $p$ depends on $\mathit{es}(p)$, what event set does $\varphihat(p)$ depend on?  We can construct a safe approximation of $\mathit{es}(\varphihat(p))$: If $\mathrm{fix}(x)\in\mathit{es}(p)$, put $\mathrm{fix}(x)\in\mathit{es}(\varphihat(p))$. For any other event $e\in\mathit{es}(p)$, put $\mathrm{dmc}(x)\in\mathit{es}(\varphihat(p))$.
This is correct because $\varphi_x$ is injective.  If $\varphi_x$ is monotone with respect to the order on $\Val_x$, $a<b \mimpl \varphi_x(a)<\varphi_x(b)$, we can also use bounds events.  If $\varphi_x$ is anti-monotone with respect to that order, we have to switch $\mathrm{lbc}$ with $\mathrm{ubc}$.

\pparagraph{Arc and path consistency}
Instead of regarding a view $\varphi$ as \emph{transforming} a constraint $c$, we can regard $\varphi$ as \emph{additional} constraints, implementing the decomposition. Assuming $\Var=\{x_1,\dots,x_n\}$, we use additional variables $x'_1,\dots,x'_n$.  Instead of $c$, we have $c'=c[x_1/x'_1,\dots,x_n/x'_n]$, which enforces the same relation as $c$, but on $x'_1\dots x'_n$. Finally, we have $n$ \emph{view constraints} $c_{\varphi,i}$, each equivalent to the relation $\varphi_i(x_i)=x'_i$. The solutions of the decomposition model, restricted to the $x_1\dots x_n$, are exactly the solutions of the original view-based model.

\begin{example}
Assume the equality constraint $c\equiv(x=y)$. In order to propagate $c'\equiv(x=y+1)$, we could use a domain complete propagator $p$ for $c$ and a view $\varphi$ with $\varphi_x(v)=v$, $\varphi_y(v)=v+1$. The alternative model would be defined with additional variables $x'$ and $y'$, a view constraint $c_{\varphi,x}$ for $x'=x$, a view constraint $c_{\varphi,y}$ for $y'-1=y$, and $c[x/x',y/y']$, yielding $x'=y'$.
\end{example}

Every view constraint $c_{\varphi,i}$ shares exactly one variable with $c$ and no variable with any other $c_{\varphi,i}$.  Thus, the constraint graph is Berge-acyclic, and we can reach a fixpoint by first propagating all the  $c_{\varphi,i}$, then propagating $c[x_1/x'_1,\dots,x_n/x'n]$, and then again propagating the $c_{\varphi,i}$. This is exactly what $\varphi^-\circ p\circ\varphi$ does. In this sense, views can be seen as a way for specifying a \emph{perfect order of propagation}, which is usually not possible in constraint programming systems.

If $\varphihat(p)$ is domain complete for $\varphi^-(c)$, then it achieves \emph{path consistency} for $c[x_1/x'_1,\dots,x_n/x'_n]$ and all the $c_{\varphi,i}$ in the decomposition model.

\section{Limitations}
\label{sec:limitations}

Although views are widely applicable, they are no silver bullet. This section explores some limitations of the presented architecture.

\pparagraph{Beyond injective views}

Views as defined in this paper are required to be injective. This excludes some interesting views, such as a view for the absolute value of a variable, or a view of a variable modulo some constant. None of the basic proofs makes use of injectivity, so non-injective views can be used to derive (bounds) complete, correct propagators.

However, event handling changes when views are not injective:
\begin{itemize}
  \item A domain change event on a variable does not necessarily translate to a domain change event on the view. For instance, given a domain $d$ with $d(x)=\{-1,0,1\}$, removing the value $-1$ from $x$ is a domain change event on $x$, but not on $\operatorname{abs}(x)$.
  \item A domain change event on a variable may result in a value event on the view. For instance, removing $0$ instead of $-1$ in the above example results in $d(x)=\{-1,1\}$, but in $\operatorname{abs}(x)$ there is only a single value left.
\end{itemize}

These effects may lead to unnecessary propagtor invocations, or even to incorrect behavior if a propagator relies on the accuracy of the reported event. As propagators in Gecode may assume that events are crisp in this sense, we decided not to allow non-injective views.

\pparagraph{Multi-variable views}
Some multi-variable views that seem interesting for practical applications do not preserve contraction, for instance a view on the sum or product of two variables. The reason is that removing a value through the view would have to result in removing a \emph{tuple} of values from the actual domain. As domains can only represent cartesian products, this is not possible in general.
For views that do not preserve contraction, Theorem~\ref{theorem:subsumtion} does not hold. That means that a propagator $p$ cannot easily detect subsumption any longer, as it would have to detect it for $\varphihat(p)$ instead of just for itself, $p$. In Gecode, propagators report whether they are subsumed, so that they are not considered for propagation again. This optimization is vital for performance, so we only allow contraction-preserving views.

For contraction-preserving views on multiple variables, all our theorems still hold. Some useful views we could identify are
\begin{itemize}
  \item A set view of Boolean variables $[b_1,\dots,b_n]$, behaving like $\setc{i}{b_i=1}$.
  \item An integer view of Boolean variables $[b_1,\dots,b_n]$, where $b_i$ is 1 iff the integer has value $i$.
  \item The inverse views of the two views above.
\end{itemize}
These views are of limited use, and the decomposition approach will probably work just as well in these cases.

\pparagraph{Propagator invariants}
Propagators typically rely on certain invariants of a variable domain implementation. If idempotence or completeness of a propagator depend on these invariants, channeling views lead to problems, as the actual variable implementation behind the view may not respect the same invariants.

For example, a propagator for interval-based finite set variables can assume that adjusting the lower bound of a variable does not affect its upper bound. If this propagator is instantiated with a channeling view for an ROBDD-based set variable, this invariant is violated: if, for instance, the current domain is $\{\{1,2\},\{3\}\}$, and you add $1$ to the lower bound, the $3$ is removed from the upper bound (in addition to $2$ being added to the lower bound). A propagator that relies on the invariant may lose idempotence.

\section{Experiments}
\label{sec:experiments}

Our experiments in~\cite{SchulteTack:Advances:2006} showed that deriving propagators using views incurs no runtime overhead. Here, we present empirical evidence for two more facts: views are highly applicable in real-world constraint programming systems, and they are clearly superior to a decomposition-based approach.

\pparagraph{Applicability}
The Gecode \CPP{} library~\cite{Gecode:2008} makes heavy use of views. Table~\ref{tab:applicability} shows the number of generic propagators implemented in Gecode, and the number of derived instances. On average, every generic propagator results in 3.59 propagator instances. Propagators in Gecode account for more than 40\,000 lines of code and documentation. As a rough estimate, generic propagators with views save around 100\,000 lines of code and documentation to be written, tested, and maintained. On the other hand, the views are implemented in less than 8\,000 lines of code, yielding a 1250\% return on investment.

\begin{table}[tb]
\caption{Applicability of views: number of generic vs.\ derived propagators}
\label{tab:applicability}       
  \footnotesize
  \centering
        \begin{tabular}{|l||r|r|r|}
        \hline
        \emph{Variable type} & \emph{Generic propagators}    & \emph{Derived propagators} & \emph{Ratio}\\
        \hline\hline
  Integer & 69 & 230 & 3.34 \\
  Boolean & 23 & 72 & 3.13 \\
  Set     & 24 & 114 & 4.75 \\
        \hline
        \emph{Overall} & 116 & 416 & 3.59\\
        \hline
        \end{tabular}
\end{table}

\pparagraph{Views vs.\ decomposition}
In order to relate derived propagators to arc and path consistency, Sect.~\ref{sec:extended_properties_of_derived_propagators} decomposed a derived propagator $\varphihat(p)$ into additional variables and propagators for the individual $\varphi_x$ and $p$. Of course, one has to ask why we advertise variable views instead of always using decomposition. Table~\ref{tab:decomposition_vs_views} shows the runtime and space requirements of several benchmarks implemented in Gecode. The numbers were obtained on a Intel Pentium IV at 2.8 GHz running Linux and Gecode 2.1.1. The figures illustrate that derived propagators clearly outperform the decomposition, both in runtime and space.

\begin{table}[tb]
\caption{Runtime and space comparison: derived propagators vs\@. decomposition}
\label{tab:decomposition_vs_views}      
\newcommand{\NOWAY}{\multicolumn{1}{c|}{---}}%
\footnotesize
  \centering
  \begin{tabular}{|l||r|r|r|r|}
    \hline%
    \emph{Benchmark} & 
    \multicolumn{2}{c|}{\emph{derived}} & 
    \multicolumn{2}{c|}{\emph{decomposed}} \\
    \cline{2-5}
    & \multicolumn{1}{c|}{\emph{time (ms)}}
    & \multicolumn{1}{c|}{\emph{space (kB)}}
    & \multicolumn{1}{c|}{\emph{relative time (\%)}}
    & \multicolumn{1}{c|}{\emph{relative space (\%)}}\\
    \hline\hline
Alpha & 91.25 & 83.22 & 405.62 & 167.32\\
Eq-20 & 1.37 & 70.03 & 613.61 & 219.95\\
Queens 100 & 24.72 & 2\,110.00 & 705.10 & 103.03\\
\hline
Golf 8-4-9 & 310.40 & 10\,502.00 & 211.47 & 231.64\\
Steiner triples 9 & 135.72 & 957.03 & 108.38 & 100.03\\
\hline
  \end{tabular}
\end{table}

\section{Conclusion and Future Work}
\label{sec:conclusion}

The paper has developed variable views as a technique to derive perfect propagator variants. Such variants are ubiquitous, and the paper has shown how to systematically derive propagators using techniques such as transformation, generalization, specialization, and channeling. 

We have presented a model of views that allowed us to prove that derived propagators are indeed perfect: they inherit correctness and domain completeness from their original propagator, and preserve bounds completeness given additional properties of views.

As witnessed by the empirical evaluation, deriving propagators saves huge amounts of code to be written and maintained in practice, and is clearly superior to decomposing constraints into additional variables and simple propagators.

For future work, it will be interesting to investigate how views can be generalized, even if that means that derived propagators are not perfect any more.

\pparagraph{Acknowledgements}

We thank Mikael Lagerkvist and Gert Smolka for fruitful discussions about views and helpful comments on a draft of this paper.

\bibliographystyle{abbrv}
\bibliography{references}

\end{document}